# Generating the Structure of a Fuzzy Rule under Uncertainty

**J.L. Castro**
Dpto. Ciencias de la Computación e I.A.
E.T.S.I. Informática. Universidad de Granada
Avenida Andalucia, 38, 18071
Granada (Spain)

**J.M. Zurita**
Dpto. Ciencias de la Computación e I.A.
Facultad de Ciencias. Universidad de Granada
Campus Fuentenueva, 18071
Granada (Spain)

## Abstract

The aim of this paper is to present a method for identifying the structure of a rule in a fuzzy model. For this purpose, an ATMS shall be used (Zurita 1994). An algorithm obtaining the identification of the structure will be suggested (Castro 1995).

The minimal structure of the rule (with respect to the number of variables that must appear in the rule) will be found by this algorithm. Furthermore, the identification parameters shall be obtained simultaneously.

The proposed method shall be applied for classification to an example. The *Iris Plant Database* shall be learnt for all three kinds of plants.

## 1 INTRODUCTION

If we want to describe a system, it is necessary to know which are the inputs and the outputs of the system, and, more importantly, the relationship between them. This function, in most cases, is not easy to achieve, and in many others, it contains highly complicated mathematical relationships.

So, it would be interesting, if this input-output conection could be obtained directly from the performance of the system. Several approaches to automatic learning have been analysed (Michalski 1984), (Quinlan 1986).

Over the last few years, the identification of systems has been carried out using fuzzy logic. This is possible since either we can not describe the system easily or for reasons of simplicity and efficiency. A set of fuzzy rules shall be worked out from a set of input-output examples of the system. This approach has been widely discussed: (De Mori 1980), (Sugeno 1991), (Campos 1993), (Delgado 1993), (González 1995).

Basically, the problem is finding a finite set of fuzzy rules able to reproduce the system's input-output behaviour. The rules shall be given in the form of:

$R$ : **if** $X_1$ is $A_1$ and ... and $X_n$ is $A_n$ **then** $Y$ is $B_i$.

As an input-output example, we shall take $E = \{e_1, \ldots, e_n, e_{n+1}\}$, where $(e_1, \ldots, e_n) \in (\mathcal{P}(D_{E1}) \times \ldots \times \mathcal{P}(D_{En}))$, and $e_{n+1} \in \mathcal{P}(D_S)$, with $\mathcal{P}$ being the Power set and $D_{Ei}$ and $D_S$ are domains of the $X_i$ and $Y$ variables, respectively.

The learning problem of a set of fuzzy rules from a set of input-output examples has two main aspects: structure identification and parameter identification. The former, tries to detect the linguistic label of the variables in the rule, and the latter gives a degree for each rule, indicating how well this rule matchies the examples.

The minimal set of variables, which must be present in the rule shall be obtained in this research. For instance, let us suppose that the following two rules had been learnt:

$R_1$ :   $X_1$ is $A \wedge X_2$ is $A' \to Y$ is $B$     [$\alpha$ uncertainty ]
$R_2$ :   $X_1$ is $A \wedge X_2$ is $A'' \to Y$ is $B'$   [$\beta$ uncertainty ],

the $X_2$ variable shall be detected as being useless in the $R_1$ rule, and so, the above rules shall be presented as:

$R'_1$ :        $X_1$ is $A \to Y$ is $B$           [$\gamma$ uncertainty ]
$R_2$ :   $X_1$ is $A \wedge X_2$ is $A'' \to Y$ is $B'$   [$\beta$ uncertainty ],

where the $R_1$ rule has been substituted by the $R'_1$ rule. Thus, the algorithm shows that, the certainty in which the $R_1$ rule matchies the whole set of examples, is included in the rule $R'_1$.

The method we propose, does not make any explicit distinction between parameter and structure identification, but both are obtained simultaneously in the learning algorithm.

The identification model idea is based on the basic principles of an ATMS (De Kleer 1986). We shall use the assumption and premise, contradiction and ATMS node terms.

Truth Maintenance Systems (TMS) are common tools in artificial intelligence for managing logical relationships between beliefs and statements. A more recently proposed ATMS (De Kleer 1986) maintains multiple sets of beliefs simultaneously, thus allowing inferences in multiple contexts at the same time. The problem-solver builds records of all the inferences mades (justifications) and hypotheses it introduces (assumptions).

The task of the ATMS is to efficiently determine, given the inferences that have been made so far, all the possible contexts and their contents. An environment is a set of assumptions. An assumption may participate in more than one environment, thus environments may overlap and even subsume one another. Therefore the set of assumptions defines an environment lattice.

Each environment defines a context. The context consists of all and only the ATMS data nodes that, given the justifications in the ATMS, can be justified entirely on the basis of the nodes in the environment. Thus, all the nodes in a given environment are also members of the derived context.

The label of each ATMS data node is a set of environments, which is required to be: Consistent: No environment in the label supports the derivation of false. Sound: For each environment in the label, the node must be included in the context defined by that environment; Complete: Any environment in the lattice whose context includes the node is a superset of some environment in the label. (Inconsistent environments have empy contexts); Minimal: No environment in the label is a subset of another.

Every ATMS node is a triple of the form $(N, L, J)$. The first element is the name of the datum being managed by the problem solver. The second element $L$ is the label. Finally, the third element is the set of justifications, $J$, having $N$ as a consequence.

## 2 DESCRIPTION OF THE APPROACH

If we have an input-output set of examples, simulating the behaviour of the system, a set of fuzzy rules and a certainty (uncertainty) value in each of them, reflecting how good is the rule with regards to the performance of the system, should be found using the learning algorithm.

The main idea is the following:

The ATMS allows the problem-solver to assume different fuzzy rules (assumptions), each of them shall be studied in relation to its identification with the set of examples taken, and, it could be rejected at any time if we find another rule which is "better" than the previous one.

The algorithm has two important aspects. Firstly, the whole environments does not have to be kept in memory, due to the ATMS minimality property, some environments (fuzzy rules), shall be removed from the label in the presence of any other one recently examined. Secondly, there is no need to examine all the rules, since, if we obtain one rule, with a matching degree to the set of examples equal to 1, we shall obviate, any other which is "subsuming" in it without even considering it.

The number of different restrictions (fuzzy linguistic labels) associated with a variable usually ranges from 5 to 9. We shall use seven, which represent, from the impossibility of the proposition to the maximun credibility in it. We can show that in Figure 1.

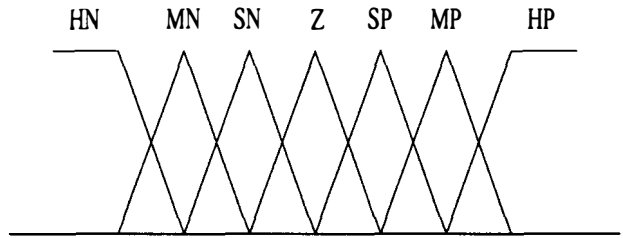

Figure 1: A set of seven linguistic labels.

Let $Y$ and $Et = \{HN, MN, SN, Z, SP, MP, HP\}$ be the output classification variable and the set of linguistic labels, respectively. Let $S$ be the Power set of $Et$, denoted as $\mathcal{P}(Et)$, this represents the set of environments which describe the ATMS lattice. Let $a$ be the size of $S$.

The learning process consists of:

All the fuzzy rules having for $Y$ both $Et_i \in Et$ as consequent and any $t \in S$ as antecedent shall be worked out. With this aim in mind, an $Et_i$ shall be established for the consequent and each $t \in S$ shall be assumed (taken as an assumption) by the problem solver. The matching degree of the rule $t \rightarrow Y_{Et_i}$, with every one of the examples $e_j, j = 1 \ldots k$ is calculated. This degree is obtained only in those examples that match (although it could be minimally) the set of predictive variables with the antecedent part of the rule. It is obvious that only these examples are considered, because the others are not identified at all by this fuzzy rule. But, of course, the latter should be identified by any other that must be found. Each $Et_i$ shall form an ATMS node $(N_{Y_{Et_i}})$, that has a matching degree greater than zero. Finally, minimality criteria shall be established among the elements of the $N_{Y_{Et_i}}$ label.

Let us look at the set of input-output examples depicted in Table 1.

When the problem solver assumes the environment $t_s \in S$, the ATMS shall update a node for $Y_{Et_i}$ made up of:

Table 1: Set of input-output examples.

|   | $X_1$ | $X_2$ | ... | $X_n$ | $Y$ |
|---|---|---|---|---|---|
| 1 | $e_{11}$ | $e_{12}$ | ... | $e_{1n}$ | $e_{Y1}$ |
| 2 | $e_{21}$ | $e_{22}$ | ... | $e_{2n}$ | $e_{Y2}$ |
|   | ......... | | | | |
| k | $e_{k1}$ | $e_{k2}$ | ... | $e_{kn}$ | $e_{Yk}$ |

$$N_{Y_{Et_i}} : ((b_1,\ldots,b_s,\ldots,b_j)(t_1,\ldots,t_s,\ldots,t_j)$$
$$(Examples_1,\ldots,Examples_s,\ldots,Examples_j)), \quad (1)$$

in which, the sequence of $b_1,\ldots,b_s,\ldots,b_j$ values means the matching degree between both the environments $t_1,\ldots,t_s,\ldots,t_j$ and the $Examples_1,\ldots,Examples_s,\ldots,Examples_j$ sets of examples, respectively. That is to say, with this node, the following sets of fuzzy rules are represented as:

$$t_s \to Y_{Et},\ [b_s];\ \forall s = 1\ldots j.$$

The $b_s$ is worked out by the problem-solver in the following way:

- *Cardinality of $t_s$ equal to $n$.* Let us supossse that $t_s$ is:

$$t_s = \{X_1 \text{ is } Et_1,\ \ldots, X_n \text{ is } Et_n\}, Et_i \in Et,$$

then, with each $\{e_1, e_2, \ldots, e_n, e_Y\}$ example, the inference:

$$\frac{X_1 \text{ is } Et_1 \quad \ldots \quad X_n \text{ is } Et_n \quad \to \quad Y \text{ is } Et_i}{e_1 \quad \ldots \quad e_n}$$
$$Y \text{ is } B'$$

shall be made.

The fuzzy set $B'$ is calculated by means of the compositional rule of inference (CRI). In this case, with crisp values, it shall be reduced to:

$$\mu_{B'}(y) = \mu_{Et_1}(e_1) \wedge \ldots \wedge \mu_{Et_n}(e_n) \wedge \mu_{Y_{Et_i}}(y).$$

- *Cardinality of $t_s$ less than $n$.* Let us suppose that $t_s$ is:

$$t_s = \{X_1 \text{ is } Et_1,\ \ldots, X_p \text{ is } Et_p\},\ Et_i \in Et,$$

Since $p < n$, then, only those variables which appear in $t_s$, shall be chosen in the example, making the inference:

$$\frac{X_1 \text{ is } Et_1 \quad \ldots \quad X_p \text{ is } Et_p \quad \to \quad Y \text{ is } Et_i}{e_1 \quad \ldots \quad e_p}$$
$$Y \text{ is } B'$$

The fuzzy set $B'$ is calculated again as:

$$\mu_{B'}(y) = \mu_{Et_1}(e_1) \wedge \ldots \wedge \mu_{Et_p}(e_p) \wedge \mu_{Y_{Et_i}}(y).$$

**Definition 2.1.**

Let $t \in S$ be an environment. Let $e$ be an input-output example. The matching degree of $t$ to $e$, that will be annotated as $MD_t(e)$, is defined as:

$$MD_t(e) = \mu_{B'}(e_Y),$$

with $B'$ being the fuzzy set obtained as mentioned in the earlier detailed inferences.

**Definition 2.2.**

Let $t \in S$ be an environment. Let $\mathcal{M}$ be a subset of the given input-output set of examples. The matching degree of $t$ to $\mathcal{M}$, that shall be annotated as $MD_\mathcal{M}(t)$, is defined as:

$$MD_\mathcal{M}(t) = Min_i\{MD_t(e_{Y_i})\}.$$

Given $t_s \in S$, $Et_i \in Et$ and the oputput variable $Y$, the learning algorithm computes:

$$t_s \to Y \text{ is } Et_i\ [MD_\mathcal{M}(t_s)].$$

Therefore the node relating to expresion (1) is:

$$N_{Y_{Et_i}} :$$
$$((MD_{Examples_1}(t_1)),\ldots, MD_{Examples_s}(t_s),\ldots,$$
$$MD_{Examples_j}(t_j))(t_1,\ldots,t_s,\ldots,t_j)$$
$$(Examples_1,\ldots, Examples_s,\ldots, Examples_j)).$$

**Definition 2.3.**

The label of a node $N_{Y_{Et_i}}$ is said to be minimal, if and only if all the environments that subsume in others are removed from the label. An environment $t_i$ subsumes in another $t_j$, if:

$$\text{subsume}(t_i, t_j) \Leftrightarrow$$
$$\begin{cases} t_j \subseteq t_i \\ MD_{Examples_j}(t_j)) > MD_{Examples_i}(t_i)) \end{cases}$$

for each $t_i, t_j \in Label(N_{Y_{Et_i}})$. That is to say, the environments giving a certainty in the proposition "$Y$ is $Et_i$" less than others (subsets of them) have, are removed.

The $t_i$ have different numbers of elements, therefore it is possible to consider $t_i$ instead of other $t_i$ which have a greater quantity of variables. When the problem-solver assumes every $t \in S$, the corresponding fuzzy rule, with respect to $t$, is worked out by the ATMS. Thus, we should establish when we can obviate a "longer" rule, and taking instead another "smaller" one. In other words, when may we say that the $\mathcal{R}$ learning fuzzy rules set is minimal?

**Definition 2.4.**

Let $\mathcal{R}$ be a learning fuzzy rules set. $\mathcal{R}$ is said to be minimal if every rule, that subsumes in others, is removed. Given two rules:

$$R_i : t_i \rightarrow Y \text{ es } Et_s \ [\alpha]$$
$$R_j : t_j \rightarrow Y \text{ es } Et_s \ [\beta]$$

$R_i$ subsumes in $R_j$ is defined as:

$$\text{subsume}(R_i, R_j) \Leftrightarrow \begin{cases} t_j \subseteq t_i \\ \beta > \alpha \end{cases}$$

**Corollary.**

$\mathcal{R}$ is minimal $\Leftrightarrow Label(N_{Y_{Et_i}})$ is minimal.

The proof is evident, taking $\alpha = MD_{Examples_j}(t_j)$ and $\beta = MD_{Examples_i}(t_i)$.

## 3   EXAMPLE

The goodness of the algorithm has been tested on the *Iris Plant Database* (Fisher 1936). This data base has been employed in many publications. Relevant information about it, may be seen in Table 2.

Table 2: Relevant Information of *Iris Plant Database*.

| | |
|---|---|
| No. Classes | 3 |
| No. Instances | 150, 50 in each class |
| No. Predictive Attributes | 4 |
| X0:Sepal Length (in cm) | Min:4.3 Max:7.9 |
| X1:Sepal Width (in cm) | Min:2.0 Max:4.4 |
| X2:Petal Length (in cm) | Min:1.0 Max:6.9 |
| X3:Sepal Width (in cm) | Min:0.1 Max:2.5 |

What we are trying to do is learn the characteristics of each plant. This will allow us to classify the four predictive attributes as belonging to one of the three classes. In order to do so, we shall use a subset, of the whole set of examples, as a training set (80 %) (120 entries), and the remaining examples shall be used as a testing set (20 %) (30 entries).

Thus, 120 examples have been chosen at random for training, and the following process has been carried out: the attributes of the iris-setosa plant have been learnt from those that are not iris-setosa, the attributes of the iris-versicolor plant have been learnt from those that are not iris-versicolor, and the attributes of the iris-virginica plant have been learnt from those that are not iris-virginica. Therefore, three fuzzy rule bases have been obtained. Each of them must be able to classify every plant.

By using the algorithm proposed, 28 of the 30 examples given in the testing set ($\frac{28}{30}$) have been classified correctly, in other words, 93.33% is the learning rate of the algorithm. What is more, 1 example cannot be distinguished as belonging to one of the three classes, therefore only 1 example of the 30 supplied is classified erroneously.

The comparison with other classic algorithms is shown in Table 3.

Table 3: Comparation of learning algorithms.

| CART | C4 | Proposed |
|---|---|---|
| 90% | 90% | 93.33% |

## 4   CONCLUSION

This paper attempts to offer an approach for identifying a fuzzy system from a set of data (input-output entries of the system). For this purpose, an ATMS was necessary. The assistance given by the ATMS (particularly because of its minimality property), allows the minimal structure of a rule to be found. So a learning algorithm has been developed.

Finally, the model shown here, has been implemented (in C language) using a SUN-4 workstation. The proofs for the latter point have been accomplished with its help.

**Acknowledgements**

The authors acknowledge the invaluable help of Professor M. Delgado. They are also grateful to Professor J.L. Verdegay and Professor E. Trillas for their valuable advice.